# USING MACHINE LEARNING TO MEASURE SELF-EFFICACY BELIEF THROUGH LEARNERS' INTERACTION TRACES IN TECHNOLOGY-ENHANCED LEARNING

Maeva Somny
*Université Lyon 1, CNRS, INSA Lyon, LIRIS, UMR5205, Villeurbanne, France*
*maeva.somny@univ-lyon1.fr*

Linda Chouati
*Université Lyon 1, CNRS, INSA Lyon, LIRIS, UMR5205, Villeurbanne, France*

Marie Lefevre
*Université Lyon 1, CNRS, INSA Lyon, LIRIS, UMR5205, Villeurbanne, France*

**ABSTRACT**

Self-efficacy beliefs play a critical role in learners' performance, academic choices, and, more broadly, in their future career trajectories. As Technology-Enhanced Learning (TEL) environments have demonstrated strong potential for supporting both performance and motivation, investigating self-efficacy within such environments is of particular interest, especially for implementing remedial practices as early as possible to support learners' self-efficacy beliefs. Our research aims to develop a system capable of detecting low self-efficacy based on learners' interaction traces while they solve exercises in a TEL system. In this paper, we present initial results and future perspectives on a detection approach based on data mining and machine learning techniques to identify features indicative of low self-efficacy beliefs. Because the detection of learners' self-efficacy beliefs through their use of TEL systems remains largely unexplored, this work is exploratory in nature. To this end, we conducted an experiment across two universities to collect interaction traces and self-reported measures of self-efficacy. We then developed a machine learning pipeline comprising multiple classification models, enabling us to highlight the importance of closely monitoring of learners' self-efficacy beliefs for improved prediction performance, given their highly dynamic nature during learning.

## 1. INTRODUCTION

Self-efficacy beliefs, defined as learners' judgments about their capability to organize and execute actions required to achieve specific goals (Bandura, 1978), play a central role in learning. A substantial body of research has shown that self-efficacy influences performance, persistence in the face of difficulties, learning strategies choice, and educational decision-making (Multon et al., 1991; Zimmerman, 2000). But self-efficacy belief is also strongly connected to motivation, anxiety or engagement (Bandura, 1997; Santi et al., 2024), some emotional states that also influences learners' behavior in an academic context. Beyond this, self-efficacy also shapes learners' long-term educational pathways and career trajectories (Ahmad & Safaria, 2013). Low self-efficacy is frequently associated with disengagement, avoidance behaviors, and reduced perseverance, negatively affecting learning outcomes (Pajares, 1996).

In parallel, Technology-Enhanced Learning (TEL) environments have become increasingly prevalent in higher education and training contexts. These environments provide learners with interactive learning activities while automatically collecting fine-grained interaction data, commonly referred to as learning traces. Such data offer valuable opportunities to analyze learners' behaviors and learning processes in situ. Prior research has demonstrated that TEL systems can effectively support other learners' traits, such as performance (Rahayu et al., 2024; Vanlehn, 2011) and motivation (Aleven et al., 2025), particularly when they incorporate adaptive mechanisms or personalized feedback (Grawemeyer et al., 2015).

In this context, the early detection of low self-efficacy represents a critical challenge. Identifying learners who experience low self-efficacy at an early stage would make it possible to deploy timely remedial or supportive interventions before difficulties become entrenched. However, despite the recognized importance of self-efficacy in learning, its automatic detection through learners' interactions with TEL systems remains largely unexplored. This paper addresses this gap by investigating the feasibility of detecting low self-efficacy beliefs from learners' interaction traces using data mining and machine learning techniques.

# 2. RELATED WORKS

## 2.1 Self-Efficacy in Learning Contexts

Self-efficacy has been extensively studied within educational psychology and the learning sciences (Bandura, 1977; Zimmerman, 2000). Research consistently shows that learners with higher self-efficacy are more motivated, engage more deeply in learning tasks, select more effective cognitive and metacognitive strategies, and achieve higher performance (P. Pintrich & Groot, 1990). Self-efficacy is also known to influence learners' persistence and their responses to failure, making it a key predictor of long-term learning success (Bandura, 1997). Research also shows that self-efficacy belief is constantly evolving, due to the learning experience : grades, teachers' feedback and behavior, classmates' success and, more broadly, all situations in which the targeted competence are put into practice (Bandura, 1977).

Traditionally, self-efficacy has been assessed using self-report questionnaires administered at specific points in time (Bandura, 2006; P. Pintrich, 2004; P. R. Pintrich et al., 1993; Zimmerman & Kitsantas, 2005). While these instruments are widely validated, they suffer from several limitations. They are intrusive, rely on learners' introspective abilities, and provide only static snapshots of a construct that is inherently dynamic and context-dependent. Other studies such as (Ba-Aoum et al., 2024) aims to predict self-efficacy belief, but only focus on sociodemographic context to achieve these predictions. Consequently, measures of self-efficacy belief are poorly suited to continuous monitoring during learning activities.

## 2.2 TEL Environments and Learning Trace Analysis

TEL environments generate detailed logs of learners' interactions, including response times, number of attempts, errors, hint usage, navigation patterns, and task completion behaviors. These interaction traces have been widely exploited in Educational Data Mining and Learning Analytics to model learners' performance (Minn & Nkambou, 2025), engagement (Cocea & Weibelzahl, 2009), affective states (Salmeron-Majadas et al., s. d.), and motivation (Boulay et al., 2007). Numerous studies have successfully leveraged machine learning methods to infer latent variables from behavioral data (Baker & Yacef, 2009).

Despite these advances, most existing work has focused on constructs such as knowledge mastery, engagement, or transient affective states (e.g., frustration or boredom) (Conati & Maclaren, 2009). The specific problem of detecting self-efficacy beliefs from interaction traces has received limited attention. This gap can be attributed to the subjective and latent nature of self-efficacy (Bandura, 1997; Pajares, 1996), as well as to its rapid evolution during learning activities (Usher & Pajares, 2008; Zimmerman, 2000).

## 2.3 Machine Learning and Dynamic Learner Modeling

Supervised Machine Learning approaches have demonstrated their effectiveness in modeling unobservable learner characteristics by linking behavioral features to ground-truth labels obtained through questionnaires or expert assessments (Baker & Yacef, 2009; Romero & Ventura, 2020). However, many existing models implicitly assume that learner characteristics remain relatively stable over a learning session or task.

This assumption is problematic in the case of self-efficacy, which can fluctuate significantly in response to task difficulty, feedback, success, or failure (Bandura, 1997; Usher & Pajares, 2008). Consequently, there is a need for models that consider the highly dynamic nature of self-efficacy, as well as for empirical investigations into how temporal monitoring influences detection performance. To date, few exploratory studies have examined how continuous or fine-grained measurement of self-efficacy affects prediction accuracy in TEL environments.

Overall, the literature highlights a lack of studies dedicated to automatically detecting low self-efficacy from learning traces and to understanding how its temporal dynamics can be leveraged to improve model performance.

# 3. METHODOLOGY

Based on the identified gaps in the literature, this study addresses the following main research question: **To what extent can learners' interaction traces within a Technology-Enhanced Learning environment be used to automatically detect low self-efficacy beliefs using Data Mining and Machine Learning techniques?** This general question is further refined into the two sub-questions:

- **(RQ1)** Which behavioral features extracted from interaction traces are most indicative of low self-efficacy beliefs?
- **(RQ2)** Does close and continuous monitoring of self-efficacy beliefs lead to more accurate predictions compared to more static or sparse measurements?

To address these research questions, we have adopted an exploratory approach, as the existing literature does not provide any formalized indicators of low self-efficacy beliefs during learners' activity, either in general or specifically within the context of TEL environments. Consequently, we constructed a dataset grounded in authentic learner interactions, collected during real learning activities, and complemented by data related to learners' academic experience.

## 3.1 Dataset

To build our dataset, and because of the exploratory nature of our work, we add **trackers all over the TEL used by the learner** to observe their interactions within the platform. These interactions, presented in Table 1 are collected while the learners are doing exercises, such as pairing exercises, grouping exercises or multiple-choice questions. These types of exercises, available on the ASKER platform (Guin & Lefevre, 2022), the TEL used by the learners, are very common and available on other more widely used platforms such as Moodle (Dougiamas & Taylor, 2003) , allowing us to replicate our approach on other TEL.

Table 1. Types of interaction traces observable within the learning platform

| INFORMATIONS | TYPE | SIGNIFICANT PARTICULARITIES |
|---|---|---|
| Common informations to all exercises types | Click on **generate exercise** button | Informations on the exercise (number of step, current step, exercise model) |
| | Click on **validate** button (inside an exercise) | |
| | Click on **return** button (inside an exercise) | |
| | Click on **try again "↺"** button (inside an exercise) | |
| | Click on a **step selection** button (inside an exercise) | |
| MCQ exercises | **Check or uncheck a proposition** in the context of a MCQ exercise | Position of the selected proposition in the propositions list, proposition number, action (check/uncheck) |
| Grouping exercises | **Drag and drop a card from a group to another** in the context of a Grouping exercise | Source and destination of the selected card |
| Pairing exercises | **Associate an answer-card to a proposition** in the context of a Pairing exercise | Source and destination of the selected card, position of the destination and source propositions in the propositions list |

In addition, we conducted several surveys throughout the semester to collect information about learners' self-efficacy beliefs. These surveys allow us to label our data. For this purpose, we selected the **Motivated Strategies for Learning Questionnaire** (MSLQ (P. R. Pintrich et al., 1993)), as one of its sections specifically focuses on self-efficacy beliefs. The MSLQ consists of two parts: the first focuses on motivation, and the second on learning strategies. Since our work focuses on self-efficacy, we restrict the MSLQ to its motivational component (as detailed later). This component includes measures of self-efficacy beliefs, as well as other highly relevant constructs such as anxiety, which is closely related to self-efficacy.

Although other data could have been collected, as physiological data which are used in other studies, we decided to feed our system only with interactions data, which allows us to have a system that is as unobtrusive as possible and the easiest to implement in the future.

## 3.2 Feature Extraction and Feature Sets

We then construct features based on TEL system usage. Given the exploratory nature of our approach, we extract all features available from the interaction dataset. These features are intended to be used as input to the machine learning pipeline in the form of feature sets. Some of these features are directly derived from interaction traces, while others are computed using custom algorithms, enabling us to characterize more complex behaviors:

- **Revision behavior** (i.e., engaging in exercises only before an exam, only after an exam, or both before and after an exam) is determined by analyzing each learner's interaction traces to identify when exercises were completed (i.e., validated, not merely generated).
- **Total number of hesitations** is computed by examining each learner's interaction traces, grouped by exercise. A hesitation in a multiple-choice question corresponds to an option being selected and deselected; in grouping or matching exercises, to a card being moved from a response position.
- **Number of hesitations by exercise type** corresponds to the total number of hesitations, considering the type of exercise.

Our feature set, presented in Table 2, is composed of features grouped according to their hypothesized meaning in an educational context.

- **Engagement**, defined as learners' willingness to participate in routine school activities, as well as their involvement in the learning environment and in extracurricular activities considered to have educational benefits (Chapman, 2003; Fletcher, 2005), comprises all features related to heavy or light use of the TEL system (feature set Table 2(a)) (Henrie et al., 2015).
- **Behavior** comprises all features related to learners' actions while completing exercises, reflecting their actions, reactions, and interactions with the learning environment (feature set Table 2(b)).
- Lastly, **Performance**, which serves as an indicator of how well learners understand and apply what they study, is characterized by features that provide information about learning outcomes (feature set Table 2(c)).

Table 2. Observable features derived from learner interactions with the platform

| **SET** | (a) ENGAGEMENT | (b) BEHAVIOR | (c) PERFORMANCE |
|---|---|---|---|
| **FEATURES** | Number of interactions | Average number of interaction (per exercise) | Average mark |
| | Total time of activity | Average time of activity (per exercise) | |
| | Number of validation | Dropout rate | |
| | Number of different exercises | Hesitation rate | |
| | Number of different exercises model | Hesitation rate by exercise types | |
| | | Retry when success rate | |
| | | Retry when fail rate | |
| | | Retry rate | |
| | | Revision behavior | |
| | | Has revised | |

Our primary focus is on the behavior feature set, to investigate whether self-efficacy beliefs can be predicted based solely on learners' behavioral data. This choice is motivated by the well-established influence of self-efficacy on both learner engagement and performance. To further examine the impact of engagement and performance on prediction results, we evaluate multiple combinations of the behavior feature set with the other feature sets.

## 3.3 Machine Learning Pipeline

The machine learning pipeline comprises a wide variety of models, all suitable for solving classification tasks, as our goal is to predict whether a learner belongs to the class of learners with low self-efficacy beliefs or to the class without low self-efficacy beliefs.

The pipeline includes classification models from different algorithmic families, enabling the evaluation of classical machine learning approaches to identify the models best suited to our features (cf. Table 3).

Since prior research highlights the critical importance of model explainability and given our requirement for a trustworthy system as well as the nature of our work (educational context, personal data, tracking of interactions), we prioritize inherently interpretable models, such as decision trees, when designing systems intended for learner use. These models offer intrinsic interpretability, as their structure allows predictions to be directly understood and explained, and they also provide a valuable means of identifying which behavioral patterns are associated with self-efficacy beliefs.

Table 33. Classification models used in the pipeline, grouped by algorithm family

| MODEL FAMILY | ALGORITHMS |
|---|---|
| Tree-based models | Decision Tree, Random Forest, Extra Trees |
| Ensemble methods | Bagging, Random Forest, Gradient Boosting, XGBoost, CatBoost |
| Linear models | Logistic Regression, Ridge Classifier |
| Distance-based methods | K-Nearest Neighbors |
| Margin-based models | Support Vector Classifier |
| Probabilistic models | Naive Bayes |
| Neural networks | Multi-Layer Perceptron (shallow, deep) |

# 4. EXPERIMENT

Our data collection experiment was conducted at Lyon1 University with 422 first-year learners enrolled in mathematics and computer science programs, during a one-semester (four-month) introductory algorithms course. Learners could use a TEL system during the course to practice exercises. Participation was voluntary, and learners could opt out of data usage in accordance with the GDPR[1].

## 4.1 Experimental Context and Data Collection

ASKER is a TEL platform that provides learners with self-assessment exercises. It supports the creation of various types of exercises, including matching, grouping, short open-ended questions, and multiple-choice questions. Learners can request the generation of multiple versions of exercises, allowing them to assess their understanding of a given concept several times (Guin & Lefevre, 2022).

The course is organized into chapters that structure the progression. For each chapter, a new set of exercises is made available on ASKER, and instructors offer optional tutoring sessions for learners who wish to attend. During the course, two exams take place: one at mid-semester and one at the end of the semester.

Over the four-month semester, the 422 learners generated approximately 14614 interaction traces on ASKER platform.

[1] The GDPR is an EU regulation that governs the processing of personal data and imposes obligations on organizations to ensure data protection and integrity.

To collect information about learners' self-efficacy beliefs, we administered **the MSLQ survey** multiple times during the semester, as one of its subscales directly measures self-efficacy beliefs. In total, five surveys were administered, corresponding to the five course chapters: **No. 1** administered one week after the start of the course, near the first support session; **Nos. 2, 3, and 4** administered during the semester, near support sessions. No. 3 took place one week before the mid-semester exam; **No. 5** administered one week before the course final exam.

## 4.2 Surveys Overview

Surveys 1 and 5 are precisely composed of the first part of MSLQ focusing of motivation, while others are composed of the sections learning belief, self-efficacy and anxiety of this first section. We reduced the survey because the study was conducted in an authentic educational setting in which participation was entirely voluntary. Learners were not required to take part in the study, nor were they obliged to complete all the administered questionnaires. To increase the likelihood of obtaining responses, we therefore limited the number of questions to those specifically related to self-efficacy beliefs and associated feelings.

Table 44. Participant evolution and self-efficacy score distributions across surveys

| | ***(a)*** **Participants (total)** | ***(b)*** **Participants with low self efficacy belief** | **Student MSLQ scores distribution,** by intervals | | | | | | | | | | | |
|---|---|---|---|---|---|---|---|---|---|---|---|---|---|---|
| | | | ***(c)*** **Self-efficacy mean distribution** | | | | | | ***(d)*** **Full survey mean distribution** | | | | | |
| | | | [1; 2] | ]2; 3] | ]3; 4] | ]4; 5] | ]5; 6] | ]6; 7] | [1; 2] | ]2; 3] | ]3; 4] | ]4; 5] | ]5; 6] | ]6; 7] |
| Survey 1 | 132 | 16 | 4 | 12 | 33 | 34 | 30 | 19 | 0 | 1 | 15 | 65 | 47 | 4 |
| Survey 2 | 79 | 7 | 2 | 5 | 16 | 27 | 14 | 15 | 1 | 4 | 16 | 36 | 19 | 3 |
| Survey 3 | 35 | 8 | 3 | 5 | 4 | 5 | 10 | 8 | 0 | 5 | 10 | 11 | 6 | 3 |
| Survey 4 | 37 | 1 | 0 | 1 | 14 | 8 | 6 | 8 | 0 | 1 | 13 | 13 | 8 | 2 |
| Survey 5 | 61 | 4 | 0 | 4 | 13 | 18 | 13 | 13 | 0 | 1 | 11 | 27 | 19 | 3 |

During the semester, the participation rate decreases over time. Table 4(a) shows that the total number of participants dropped from 132 to 61 between Surveys 1 and 5. The distribution of learners between those with low self-efficacy beliefs and those without is strongly skewed toward learners not exhibiting low self-efficacy beliefs, as we can see on Table 4(b). Low self-efficacy beliefs are determined based on the learners' mean scores on the self-efficacy section of the questionnaire. Learner with a mean score lower than 3/7 are considered to have low self-efficacy for this descriptive analysis. Table 4(c) presents a finer-grained distribution of learners' mean self-efficacy score, while Table 4(d) shows the distribution of mean scores across the entire survey.

Figure 15. Participant evolution and self-efficacy score distributions across questionnaires. 10 students per figure. The red dashed line marks the mid-term exam

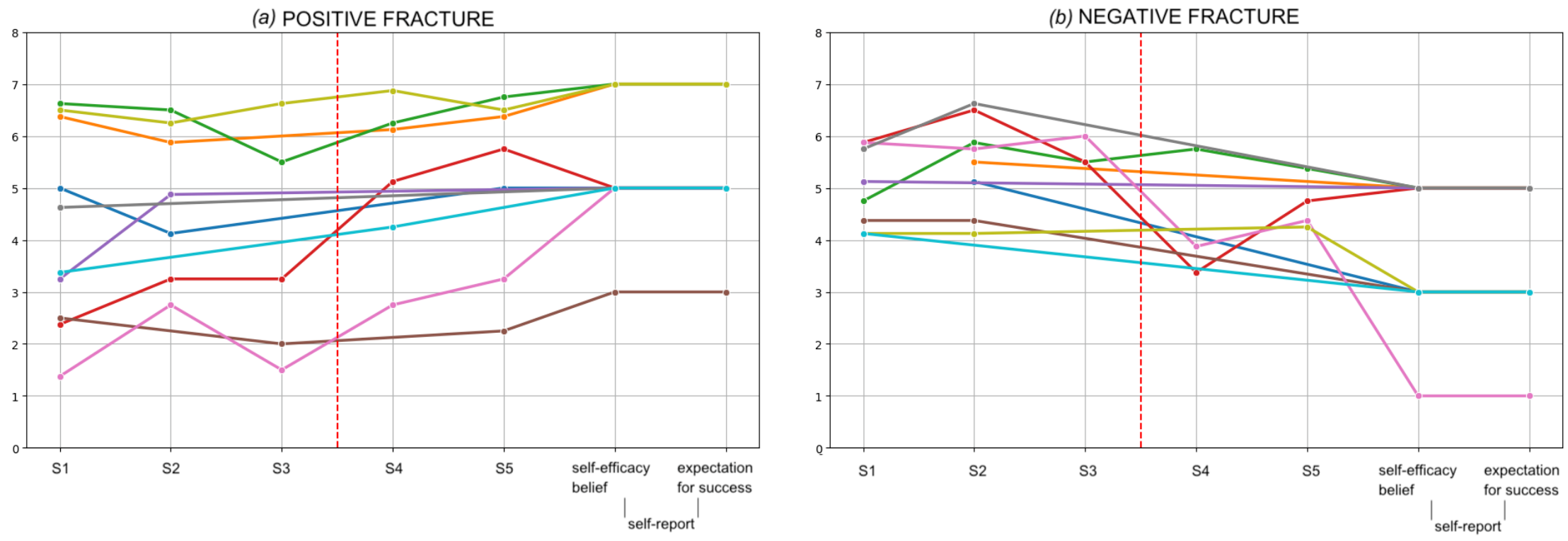

The surveys allow us to identify important events that may affect learners' self-efficacy beliefs, such as the mid-term exam. This exam appears to have an impact on the self-efficacy beliefs of some learners, as illustrated in Figure 1. The figure shows two types of learners, based on the variation of the scores obtained to MSLQ (10 learners per graph for example) and highlights a clear shift occurring between Surveys 3 and 4 in some cases, corresponding to the mid-term exam and the release of grades. Figure 1(a) shows a positive effect of the exam on self-efficacy belief, with a significant increase of the students' scores between survey 3 and 4. On the other hand, Figure 1(b) shows a negative effect of the exam on the self-efficacy belief of the student represented in the graphic. This overview still needs deeper analysis to be confirmed, by comparing survey results to real students' competencies, and bringing out if these evolutions are due to a students' awareness of their difficulties, or not.

Figure 2 presents a more in-depth analysis of the survey data, allowing that several sections of the MSLQ closely reflect the self-efficacy belief dimension. These figures are calculated based on every answer since we want to have an overview of every student, even if there is a strong disparity in learners' participation to survey. Keeping only the student who answers to every survey may insert a bias on this result since they are also (in the majority) the most engage student. Once again, we can observe the mid-term exam shift in the boxplots of the "self-efficacy", especially when looking at the median evolution (higher self-efficacy before the exam than after), and "anxiety", which is higher (lower score) just before the exam.

Figure 26. Boxplots illustrating the distribution of scores for the different sections of the MSLQ questionnaire across survey administrations

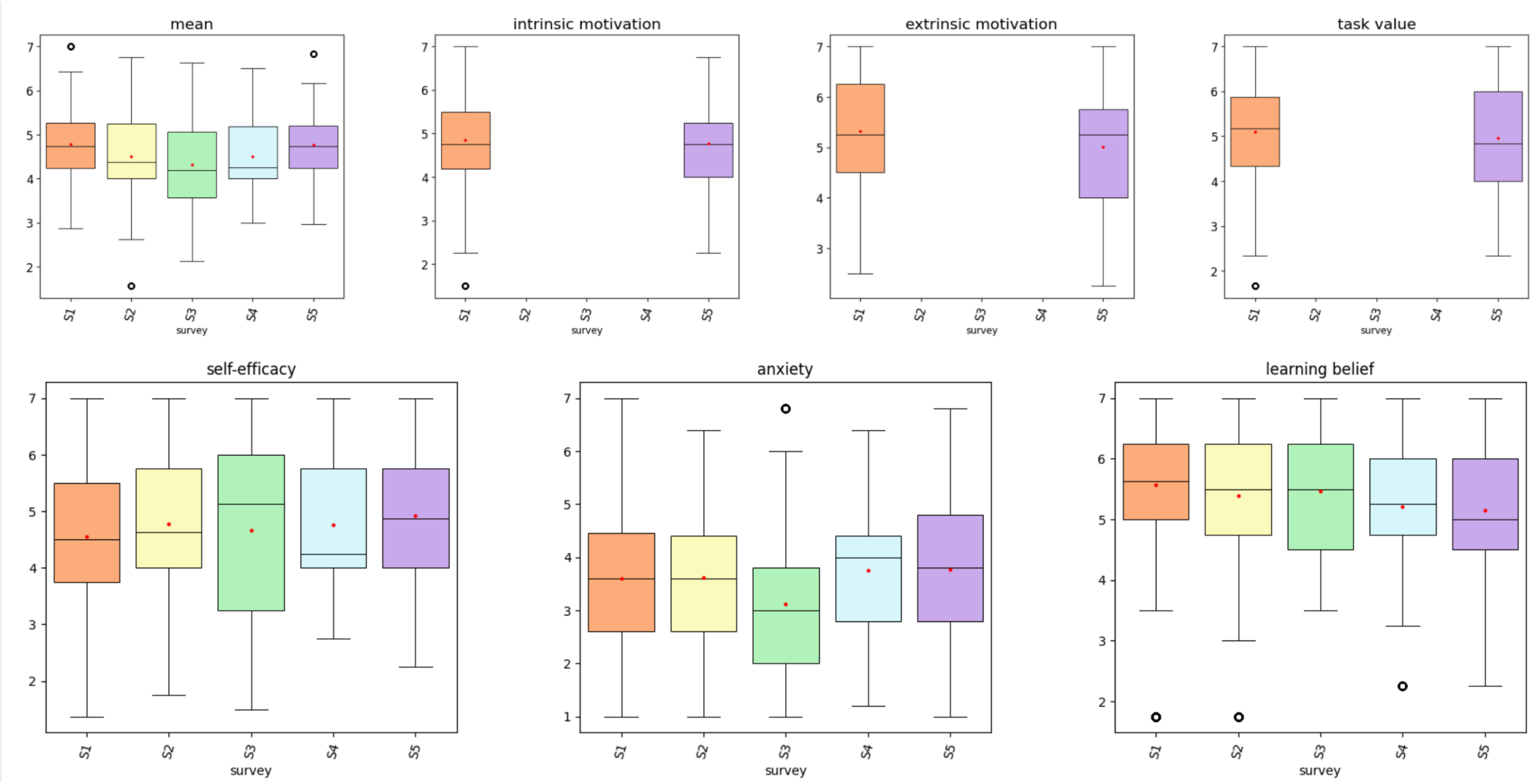

This descriptive analysis of our data indicates that questionnaires are limited tools for the purposes of our study. First, as the total number of participants decrease over the course of the semester, we observe a corresponding decline in the proportion of learners exhibiting low self-efficacy beliefs. This trend may be explained by the dropout or disengagement among learners with low self-efficacy, echoing findings reported in prior research on the self-efficacy beliefs. Since we are specifically interested in behaviors that may indicate low self-efficacy, we require data that are reliably labeled as such.

Second, research has shown that self-efficacy beliefs evolve over time. The decline in participation highlights the subgroup of learners who responded to all surveys (only 17 learners out of the 422-learner sample), enabling the most precise labelling of data. As the mid-term exam appears to affect the self-efficacy beliefs of some learners, we can also label data based on measurements collected before and after this event. The number of learners who completed at least one survey before, and at least one survey after the mid-term exam is larger (146 learners), allowing for a broader analysis.

Finally, although surveys allow us to label data at the weekly level, self-efficacy beliefs may evolve much more rapidly. It would therefore be beneficial to find methods that enable finer-grained data labeling while being less burdensome for learners, thereby increasing both participation rates and temporal resolution.

## 5. RESULTS

**The structure of our pipeline allows us to evaluate multiple configurations** for the automated application of different models to various feature sets (behavior only, behavior + performance, behavior + engagement, and behavior + engagement + performance), under several labeling strategies. This design enables us to test the following configurations:

- **Time-period-based labeling**: Labeling can be configured by splitting the dataset according to the mid-term exam or the five survey administrations. After splitting the data, features are computed for each time segment and labeled using the corresponding questionnaires.
- **MSLQ-based labeling**: Labeling can be configured by considering either the full questionnaire or only the section focusing on self-efficacy beliefs.
- **Low self-efficacy threshold**: Labeling can be configured by adjusting the cutoff score required to assign instance to one class or the other.

Table 5 presents on overview of the results obtained across the different configurations. For each configuration, only the three models achieving the highest accuracy are reported, provided that their p-value is below 0.2. A distinction is made between the F1-scores of class 0 (learners with low self-efficacy beliefs) and class 1 (learners without low self-efficacy beliefs) due to the strong class imbalance, making the F1-score the primary indicator of model performance. The results reveal a clear imbalance between the two F1-score, in favor of class 0. This effect may be explained by the way the classes are defined; consequently, several threshold configurations must be considered in order to better balance these performance indicators.

Table 57. Results across pipeline configurations. Label 0 denotes low self-efficacy and Label 1 non-low self-efficacy (threshold: mean ≤ 3/7). Models with a p-value ≤ 0.1 are highlighted in bold (** p ≤ 0.01; * 0{,}01< p≤ 0.05)

| | SURVEY SPLIT | | | | | | | | | | MID-TERM EXAM SPLIT |
|---|---|---|---|---|---|---|---|---|---|---|---|
| | All MSLQ sections used for labelisation | | | | | Self-efficacy section of MSLQ used for labelisation | | | | | |
| **Feature set** | **Top Model** | Class 0 F1 score | Class 1 F1 Score | Accuracy | p-value | **Model** | Class 0 F1 score | Class 1 F1 Score | Accuracy | p-value | |
| Behavior | SVC | 0.655 | 0.375 | 0.556 | 0.178 | MLP (1 layer) | 0.787 | 0.552 | 0.711 | **0.046*** | No significant p-value |
| | MLP (3 layers) | 0.6 | 0.5 | 0.556 | **0.042*** | KNN | 0.769 | 0.4 | 0.667 | **0.016*** | |
| | Gaussian Naive Bayes | 0.631 | 0.364 | 0.533 | 0.16 | MLP (3 layers) | 0.746 | 0.516 | 0.667 | **0.088** | |
| Behavior + engagement | KNN | 0.724 | 0.5 | 0.64 | **0.01**** | KNN | 0.767 | 0.533 | 0.689 | **0.037*** | |
| | Bagging | 0.71 | 0.357 | 0.12 | 0.129 | ExtraTrees | 0.754 | 0.483 | 0.667 | 0.102 | |
| | Logistic Regression | 0.678 | 0.387 | 0.57 | **0.018*** | CatBoost | 0.75 | 0.385 | 0.644 | **0.094** | |
| Behavior + performance | Decision Tree | 0.721 | 0.413 | 0.622 | **0.046** | KNN | 0.762 | 0.4 | 0.667 | **0.033*** | |
| | Gaussian Naive Bayes | 0.643 | 0.411 | 0.556 | **0.092** | ExtraTrees | 0.758 | 0.333 | 0.644 | 0.186 | |
| | CatBoost | 0.582 | 0.343 | 0.489 | **0.109*** | CatBoost | 0.721 | 0.414 | 0.622 | 0.141 | |
| Behavior + engagement + performance | Logistic Regression | 0.607 | 0.353 | 0.511 | **0.004**** | CatBoost | 0.781 | 0.461 | 0.689 | 0.136 | |
| | XG Boost | 0.582 | 0.343 | 0.489 | 0.151 | KNN | 0.754 | 0.483 | 0.667 | **0.041*** | |
| | Ridge | 0.582 | 0.343 | 0.489 | **0.012*** | Ridge | 0.733 | 0.467 | 0.644 | **0.076** | |

Indeed, accuracy can be misleadingly high in certain configurations where very few instances belong to class 0, as predicting all samples as class 1 yields few errors (i.e, few class 0 instances misclassified as class 1). Despite this limitation, the observed accuracy scores remain promising, especially considering that the models rely on basic architectures with hyperparameters tuned using GridSearch.

Furthermore, as self-efficacy beliefs constitute a personal construct that may manifest differently across learners, obtaining initial prediction results that highlight common behavioral features is encouraging. In addition, given the complexity of the data, that come from real learners and their interactions, these preliminary findings demonstrate the feasibility of predicting self-efficacy beliefs from TEL usage.

Finally, the p-values reported in Table 5, computed using permutation tests combined with cross-validation, assess whether model performance is significantly better than chance.

**Configurations based on survey-level splits show better results than those based on the mid-term exam split.** These findings highlight the dynamic nature of the self-efficacy belief and provides an answer to RQ2, indicating that close and continuous monitoring is necessary to accurately predict self-efficacy. In this configuration, features are computed and labeled in a more fine-grained manner.

**Labelling based solely on the self-efficacy section of the questionnaire proves sufficient to obtain meaningful results**. The best results, in terms of both accuracy and p-value, are highlighted in yellow. Overall, performance improves when only the self-efficacy section is used for labeling. We therefore adopt this labeling strategy, as it allows us the administration of shorter surveys with fewer questions, which are more likely to be completed by learners, without sacrificing labeling precision.

**Behavior feature set, composed exclusively of learners' actions and behaviors within the TEL environment, shows strong results.** Combining this feature set with engagement and performance features can even lead to reduced model accuracy in some configurations. The red line highlights the best results among all tested combinations. These findings illustrate the relationships between self-efficacy beliefs, engagement, and performance. Moreover, since our objective is to design a prediction system that does not explicitly rely on learners' perceived competence or study diligence, both of which may themselves be influenced by self-efficacy beliefs, the results appear particularly encouraging.

Table 68. Results for the best-performing configuration after parameter tuning. *Label 0* denotes low self-efficacy and *Label 1* non-low self-efficacy. Models with a *p-value* ≤ 0.1 are highlighted in bold (** p≤ 0.01; * 0{,}01 < p ≤ 0.05)

| MODEL FAMILY | MODEL | *(C1)* 0 ≤ **2,5/7** < 1 | | | | *(C2)* 0 ≤ **3/7 < 1** | | | | *(C3)* 0 ≤ **3,5/7** < 1 | | | | *(C4)* 0 ≤ **4/7** < 1 | | | |
|---|---|---|---|---|---|---|---|---|---|---|---|---|---|---|---|---|---|
| | | Class 0 F1 score | Class 1 F1 Score | Accuracy | p-value | Class 0 F1 score | Class 1 F1 Score | Accuracy | p-value | Class 0 F1 score | Class 1 F1 Score | Accuracy | p-value | Class 0 F1 score | Class 1 F1 Score | Accuracy | p-value |
| Tree based | Decision Tree | 0.6333 | 0.2667 | 0.5111 | **0.037*** | 0.6545 | 0.4571 | 0.5778 | 0.1718 | 0.6429 | 0.4118 | 0.5556 | 0.2597 | 0.6875 | 0.2308 | 0.5556 | **0.004**** |
| | Random Forest | 0.5283 | 0.3243 | 0.4444 | **0.011*** | 0.6667 | 0.3333 | 0.5556 | 0.1089 | 0.7077 | 0.24 | 0.5778 | 0.1209 | 0.6984 | 0.2963 | 0.5778 | **0.031*** |
| | Extra Trees | 0.5385 | 0.3684 | 0.4667 | **0.036*** | 0.7188 | 0.3077 | 0.6 | **0.0819** | 0.697 | 0.1667 | 0.5556 | 0.1588 | 0.7059 | 0.0909 | 0.5556 | 0.1379 |
| Ensemble | Bagging | 0.5714 | 0.2941 | 0.4667 | **0.04*** | 0.7213 | 0.4138 | 0.6222 | 0.3916 | 0.6452 | 0.2143 | 0.5111 | **0.027*** | 0.7273 | 0.25 | 0.6 | 0.5974 |
| | Gradient Boosting | 0.5455 | 0.2857 | 0.4444 | **0.03*** | 0.566 | 0.3784 | 0.4889 | 0.1469 | 0.7213 | 0.4138 | 0.6222 | 0.1518 | 0.7647 | 0.2727 | 0.6444 | 0.1349 |
| | CatBoost | 0.5455 | 0.2857 | 0.4444 | **0.0519** | 0.6552 | 0.375 | 0.5556 | **0.0649** | 0.6333 | 0.2667 | 0.5111 | 0.1019 | 0.6875 | 0.2308 | 0.5556 | 0.2398 |
| | XGBoost | 0.4490 | 0.3415 | 0.4 | 0.1219 | 0.6038 | 0.4324 | 0.5333 | 0.2418 | 0.6441 | 0.3226 | 0.5333 | 0.3477 | 0.6765 | 0 | 0.5111 | **0.042*** |
| Linear | Logistic Regression | 0.6316 | 0.3636 | 0.5333 | **0.011*** | 0.7213 | 0.4138 | 0.6222 | **0.0939** | 0.6667 | 0.3333 | 0.5556 | **0.049*** | 0.8169 | 0.3158 | 0.7111 | **0.0719** |
| | Ridge | 0.6316 | 0.3636 | 0.5333 | **0.015*** | 0.6984 | 0.2963 | 0.5778 | 0.1369 | 0.7097 | 0.3571 | 0.6 | **0.008**** | 0.8169 | 0.3158 | 0.7111 | 0.1289 |
| Distance based | KNN | 0.6038 | 0.4324 | 0.5333 | **0.001**** | 0.7692 | 0.4 | 0.6667 | **0.015** | 0.7576 | 0.3333 | 0.6444 | **0.001**** | 0.7353 | 0.1818 | 0.6 | **0.033*** |
| Margin based | SVC | 0.5882 | 0.4615 | 0.5333 | **0.004**** | 0.7586 | 0.5625 | 0.6889 | **0.009**** | 0.6667 | 0.3333 | 0.5556 | **0.037*** | 0.6452 | 0.2143 | 0.5111 | 0.3147 |
| Probabilistic | Gaussian Naive Bayes | 0.5385 | 0.3684 | 0.4667 | **0.002**** | 0.7143 | 0.5294 | 0.6444 | 0.1269 | 0.6897 | 0.4375 | 0.6 | **0.012*** | 0.7761 | 0.3478 | 0.6667 | **0.0629** |
| MLP (1 layer) | Adam Solver | 0.5882 | 0.4615 | 0.5333 | **0.005**** | 0.7541 | 0.4828 | 0.6667 | **0.0569** | 0.7742 | 0.5 | 0.6889 | **0.0549** | 0 | 0.5 | 0.3333 | 0.3606 |
| | LBFGS Solver | 0.5769 | 0.4211 | 0.5111 | **0.011*** | 0.7667 | 0.5333 | 0.6889 | 0.1848 | 0.678 | 0.3871 | 0.5778 | 0.3457 | 0.6774 | 0.2857 | 0.5556 | 0.1219 |
| | SGD Solver | 0.6667 | 0.4242 | 0.5778 | **0.01**** | 0 | 0.5714 | 0.4 | 1 | 0.383 | 0.3256 | 0.3556 | **0.0619** | 0.2286 | 0.5091 | 0.4 | 0.2228 |
| MLP (3 layers) | Adam Solver | 0.5769 | 0.4211 | 0.5111 | **0.002**** | 0 | 0.5714 | 0.4 | 1 | 0.5714 | 0.4878 | 0.5333 | **0.0579** | 0.7606 | 0.1053 | 0.6222 | 0.1469 |
| | LBFGS Solver | 0.5200 | 0.4 | 0.4667 | **0.028*** | 0.7541 | 0.4828 | 0.6667 | 0.2647 | 0.6071 | 0.3529 | 0.5111 | 0.1748 | 0.6333 | 0.2667 | 0.5111 | 0.2088 |
| | SGD Solver | 0.5490 | 0.4103 | 0.4889 | **0.004**** | 0 | 0.5714 | 0.4 | 1 | 0 | 0.5484 | 0.3778 | 1 | 0 | 0.5 | 0.3333 | 1 |

Table 6 presents a detailed analysis of results obtained using a survey-based split, labeling based on the self-efficacy section, and the behavior feature set, as this combination yielded the best overall performance. Because previous results did not clearly identify a single model that would truly be better than the others, we conducted an extensive hyperparameter tuning with an extent GridSearch. The goal was to improve models' accuracy without making the p-value and distinct F1-scores drop.

The results indicate that modifying the threshold used for class assignments have an impact on classes F1-scores. However, achieving well-balanced F1-scores remains challenging. As the thresholds grows, the number of data labelled as 0 grows, but the class gets more diverse too, because it includes learner with "not that low" self-efficacy belief, this seems to affect p-value too. Given the available data, determining an optimal threshold remains difficult, as threshold selection directly influences class distribution. And currently, no model produces F1-scores that are balanced and close to 1.

Despite extensive tuning, accuracy remains similar to previously reported results. Notably, linear models achieve the highest accuracy during training (highlighted in green), which is encouraging given our objective of favoring explainable models. But we must keep in mind that these results also show non-relevant p-value. KNN and SVC (our most balanced model for a 3/7 threshold, highlighted in red) performs well too, when we consider models with good accuracy and relevant p-value, another advantage for XAI.

In the specific case of neural networks models, they do not perform better than others. This constitutes an additional encouraging result, both from an explainability perspective and in terms of computational and temporal efficiency, as we aim to deploy models that remain practical and scalable in real-world educational setting involving hundreds of learners.

Figure 39. Most significant features contributing to model decisions

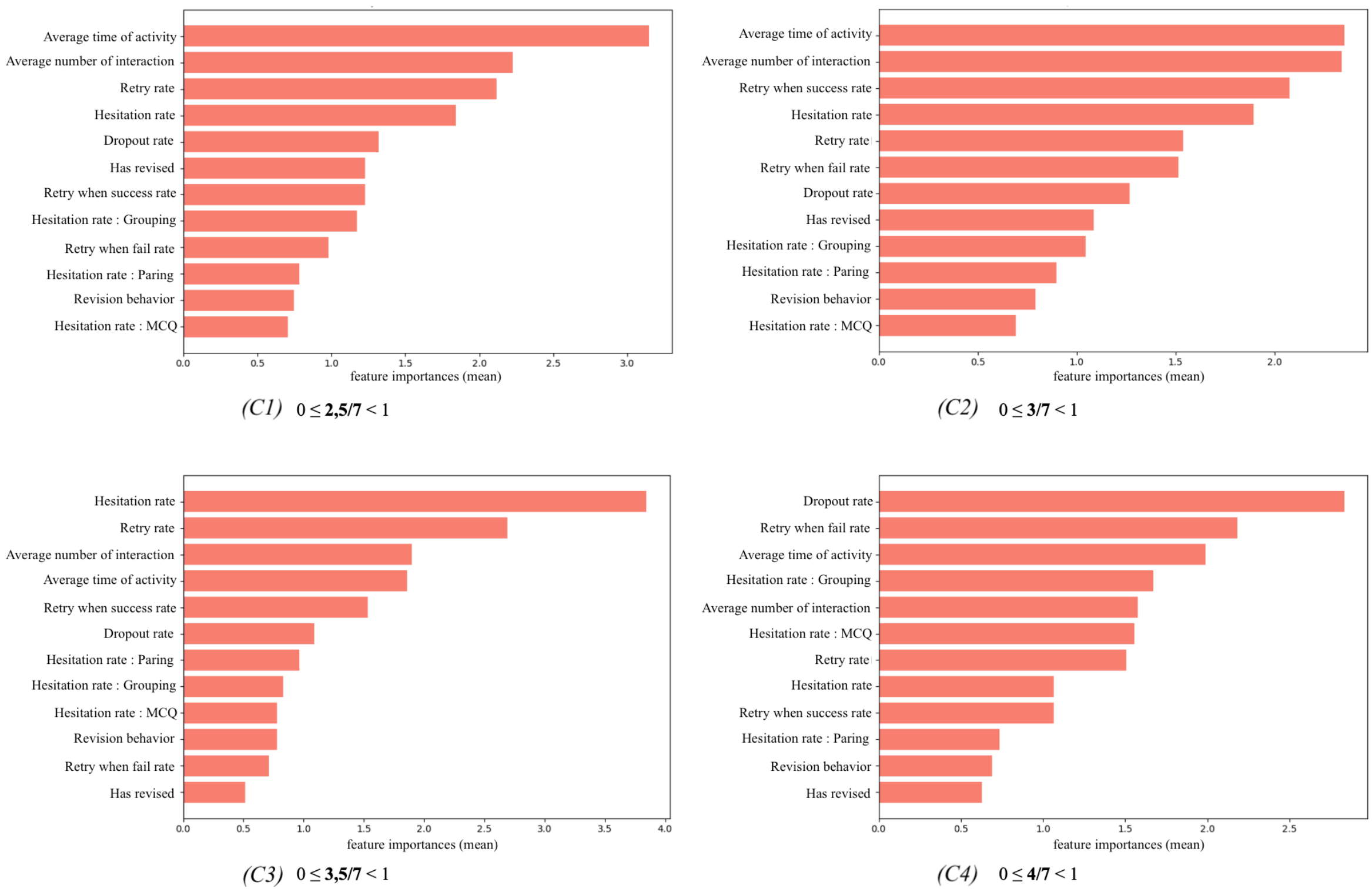

**Finally, the intrinsic interpretability of ensemble models allows us to analyse which features are the most influential factors in the model decisions.** These explanations are obtained by examining the most significant features contributing to the predictions and are affected by the chosen threshold configurations.

Figure 3 presents the most significant features for the four thresholds configurations, providing insights relevant to RQ1. The impact of threshold selection is particularly interesting to observe since a bigger threshold means learner with "not that low" self-efficacy belief. Some features, very important for the $0 \leq 4/7 \leq 1$ configuration but not particularly important in other configurations, like "dropout rate" or "retry when fail" could be some distinction between learner with low self-efficacy beliefs. The results suggest the existence of two groups among the four configurations: $0 \leq 2/7 \leq 1$ and $0 \leq 2.5/7 \leq 1$, which share very similar top-three features that are less influential in the other two configurations. This observation may help to determine the best thresholds, with the help of teacher expertise for example, since features like "retry when success" seems to transcribe a lack of self-confidence.

Configuration $0 \leq 4/7 \leq 1$ have the particularity of having "retry when fail" on the top of the list, seeming to mark a separation from the other configurations, and thus serving more as a marker of non-low self-efficacy beliefs. Given that the MSLQ survey asks learners to self-assess on a 1-to-7 scale, a threshold of four corresponds to a value above the scale midpoint. Consequently, these differences in feature importance between configurations represent a promising step toward more accurate detection of low self-efficacy beliefs.

# 6. DISCUSSION

Despite the large volume of collected interaction traces, this study is constrained by a limited amount of labeled data. This imbalance leads to overfitting in the predictive models. The main cause is the small number of completed questionnaires, which already limits the feasibility of continuous self-efficacy monitoring. This limitation is particularly critical, as self-efficacy is a dynamic construct that may evolve rapidly during learning activities.

These observations motivate the exploration of alternative labeling strategies that are less intrusive and more suitable for fine-grained and continuous monitoring.

Building on previous studies that leverage self-reported self-efficacy measurements (Bernacki et al., 2015), one possible direction would be to prompt learners to self-declare their perceived self-efficacy at specific moments of interaction, for instance when completing exercises. Such an approach could enable more frequent and temporally aligned self-efficacy annotations, thereby supporting finer-grained labeling while reducing the reliance on episodic questionnaires. However, caution is required to avoid overburdening learners with frequent prompts, as excessive solicitation may lead to response fatigue and may compromise the accuracy of self-reported data (Bandura, 1997; Shiffman et al., 2008). Careful design of the frequency, timing, and intrusiveness of such self-declarations is therefore essential to balance data quality with learner experience.

Although the current results are not sufficient to support the deployment of a reliable detection tool for learner support, the study remains promising. It contributes to a broader line of research on the prediction of perceived competence and addresses a gap in the existing literature. Furthermore, despite the personal and heterogeneous nature of self-efficacy beliefs, our results indicate that common behavioral features can be identified to enable their prediction from learner interactions in TEL environments.

# ACKNOWLEDGEMENT

This work is supported by TeaTime Project (ANR-23-CE38-0012) & ASLAN Labex (ANR-10-LABX-0081).